\documentclass{article}

\usepackage{arxiv}

\usepackage[utf8]{inputenc} 
\usepackage[T1]{fontenc}    
\usepackage{hyperref}       
\usepackage{url}            
\usepackage{booktabs}       
\usepackage{amsfonts}       
\usepackage{nicefrac}       
\usepackage{microtype}      
\usepackage{lipsum}
\usepackage{graphicx}
\graphicspath{ {./images/} }
\usepackage{amsmath}   
\usepackage{amssymb}
\usepackage{algorithm}
\usepackage{algpseudocode}
\usepackage{bm}
\newcommand{\vect}[1]{\bm{#1}}

\title{Prototype-Driven Adaptation for Few-Shot Object Detection}

\author{
 Yushen Huang \\
  School of Coumputer and Communication Engineering\\
  University of Science and Technology Beijing\\
  Beijing, China  \\
  \texttt{2498844026@qq.com} \\
  \And
 Zhiming Wang \\
  School of Coumputer and Communication Engineering\\
  University of Science and Technology Beijing\\
  Beijing, China \\
  \texttt{wangzhiming@ustb.edu.cn} \\
}

\begin{document}
\maketitle
\begin{abstract}
Few-shot object detection (FSOD) often suffers from base-class bias and unstable calibration when only a few novel samples are available. We propose Prototype-Driven Alignment (PDA), a lightweight, plug-in metric head for DeFRCN that provides a prototype-based “second opinion” complementary to the linear classifier. PDA maintains support-only prototypes in a learnable identity-initialized projection space and optionally applies prototype-conditioned RoI alignment to reduce geometric mismatch. During fine-tuning, prototypes can be adapted via exponential moving average(EMA) updates on labeled foreground RoIs—without introducing class-specific parameters—and are frozen at inference to ensure strict protocol compliance. PDA employs a best-of-K matching scheme to capture intra-class multi-modality and temperature-scaled fusion to combine metric similarities with detector logits. Experiments on VOC FSOD and GFSOD benchmarks show that PDA consistently improves novel-class performance with minimal impact on base classes and negligible computational overhead. 
\end{abstract}


\section{Introduction}

\subsection{Motivation}

Under few-shot supervision, the detector’s linear classifier tends to overfit base classes while producing biased scores for novel ones. 
DeFRCN alleviates this issue via Gradient Decoupled Learning (GDL) and static Prototypical Calibration Blocks (PCB), yet static prototypes derived from external features may \textbf{misalign with the detector’s feature space} and \textbf{fail to model intra-class diversity}.


\subsection{Our Approach}

We introduce \textbf{Prototype-Driven Alignment (PDA)}---a class-agnostic, plug-in metric head that complements the classifier in DeFRCN. 
PDA initializes prototypes \textbf{solely from the support set}, maintains a \textbf{task-adaptive prototype memory}, and learns a \textbf{projection initialized as identity} to ensure stability and interpretability.
An optional \textbf{prototype-conditioned RoI alignment} further mitigates geometric mismatch between support and query RoIs.
To handle intra-class multi-modality, PDA adopts a \textbf{best-of-K similarity scheme}, and re-ranks predictions by \textbf{temperature-scaled fusion} of metric and classification logits.


\subsection{Protocol Compliance}

A key concern in FSOD is \textbf{information leakage} from query/test data.
PDA ensures compliance through three principles:
\begin{enumerate}
    \item \textbf{Support-only initialization:} prototypes are built exclusively from support features.
    \item \textbf{Optional EMA updates:} when enabled, prototypes are updated only using \textbf{labeled foreground RoIs} during fine-tuning---without any per-class learnable parameters.
    \item \textbf{Frozen inference:} prototype memory is fixed at test time.
\end{enumerate}


\subsection{Contributions}

Our main contributions are summarized as follows:

\begin{itemize}
    \item \textbf{Generalized calibration head:} 
    We extend the Prototypical Calibration Block (PCB) into a \textbf{task-adaptive, prototype-driven alignment head} that incorporates \textbf{best-of-K prototypes}, a \textbf{learnable identity-initialized projection}, and optional \textbf{prototype-conditioned RoI alignment} for geometric consistency.
    \item \textbf{Consistent empirical gains:}
    Extensive experiments on \textbf{VOC FSOD and GFSOD} benchmarks demonstrate that PDA achieves competitive performance with minimal additional parameters and negligible runtime overhead.
\end{itemize}

\section{Related Work}
\label{sec:related}

\paragraph{FSOD/GFSOD: Fine-tuning and Meta-learning.}
Few-shot object detection (FSOD) and generalized FSOD (GFSOD) are often approached by either fine-tuning or meta-learning. Fine-tuning decouples representation learning from category adaptation by first pretraining on base classes and then adapting a lightweight head (or adapter) with few novel samples; representative methods include TFA and its variants (e.g., contrastive regularization) and DeFRCN which employs gradient decoupling and an offline Prototypical Calibration Block (PCB) to mitigate multi-task interference and base bias~\cite{wang2020frustratingly,sun2021fsce,qiao2021defrcn}. Meta-learning instead trains class-agnostic adaptation with support--query episodes (e.g., feature reweighting, Meta R-CNN, MetaDet, attention-guided proposals)~\cite{kang2019few,yan2019meta,fan2020few}. Our PDA follows the standard fine-tuning pipeline and is a drop-in head for DeFRCN. It introduces a prototype-based inductive bias characteristic in metric-learning and meta-learning, but trains in a non-episodic manner.Consequently,it complements the detector by providing a metric perspective that supplements and calibrates its decisions.

\paragraph{Prototype-based Metric Learning for Detection.}
Prototype-based classifiers (e.g., ProtoNet, cosine-style heads, relation heads) decide by distances/similarities to class prototypes with few per-class parameters~\cite{snell2017prototypical,wang2018cosface,sung2018learning}. In two-stage detection, however, naively fusing metric scores with detector logits can underperform due to feature-space mismatch between support prototypes and query RoIs, as well as geometric misalignment among RoIs. Real categories are often multi-modal, motivating multi-prototype schemes (e.g., best-of-$K$) for compact yet expressive representations. PDA emphasizes alignment-aware metric learning: an identity-initialized projection for stable coupling to the current detector features, and per-class best-of-$K$ prototypes to account for intra-class diversity.

\paragraph{Calibration and Bias Mitigation in Detection.}
Score calibration is widely used to counter base-class bias; in DeFRCN, PCB fuses prototype similarity with classifier scores but remains offline and tied to a fixed classification backbone, which may be suboptimal once the detector’s feature geometry changes during fine-tuning~\cite{qiao2021defrcn}. Stability techniques such as temperature scaling and learnable scaling/bias terms can further improve fused-score reliability~\cite{guo2017calibration}. PDA retains the ``second-opinion'' idea while learning a task-adaptive metric space coupled to the current detector features, and stabilizes fusion via temperature scaling together with a learnable metric scale and background bias, without modifying detection losses.

\section{Method}
\label{sec:method}

\subsection{Design Motivation: From Static PCB to Task-Adaptive PDA}
\label{subsec:motivation}
\textbf{Problem.}
DeFRCN's Prototypical Calibration Block (PCB)~\cite{qiao2021defrcn} calibrates detector scores with
\emph{offline, frozen} features, which are not co-adapted to the geometry of detection RoIs
(occlusion, pose, scale, local background). This yields a feature-space mismatch. Coupled with a
single prototype and no geometric alignment, it may \emph{over-score} shape-similar negatives and
\emph{under-score} misaligned true positives.

\textbf{Principle.}
We introduce \emph{Prototype-Driven Adaptation} (PDA), a class-agnostic, task-adaptive metric head that:
(i) learns a shared, identity-initialized projection to embed RoIs in a stable metric space;
(ii) maintains per-class multi-prototypes with EMA updates from labeled foreground RoIs during fine-tuning;
(iii) aligns RoI maps conditioned on prototypes to reduce geometric mismatch; and
(iv) fuses metric and detector logits using temperature, learnable scale, and background bias,
all without modifying the detection losses.

\begin{figure}
  \centering
  \includegraphics[width=\linewidth]{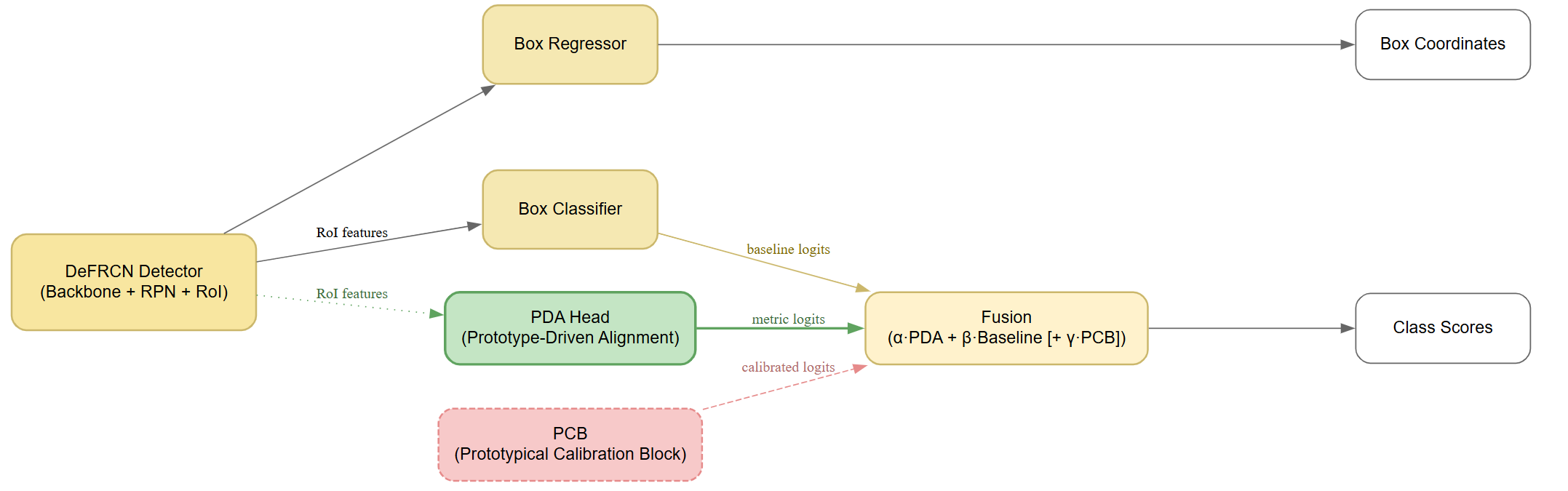}
  \caption{Overall architecture and fusion. PDA produces metric logits from projected RoI features and fuses them with baseline and PCB logits as in Eq.~\eqref{eq:fusion-vector}. Box regression is unchanged.}
  \label{fig:arch-fusion}
\end{figure}

\subsection{Dynamic Prototype Memory}
\label{subsec:wta-ema}

For each category $c \in \{1,\dots,C\}$, we maintain a $K$-slot prototype memory
$\mathcal{P}_c=\{p_{c,1},\dots,p_{c,K}\}$, $p_{c,k}\in\mathbb{R}^D$,
where $D$ is the feature dimension.
The memory is initialized from the support set:
we first extract L2-normalized ROI vectors $\{z_i\}_{i\in\mathcal{S}_c}$ for class $c$,
compute their mean $\bar{z}_c = \frac{1}{|\mathcal{S}_c|}\sum_{i\in\mathcal{S}_c} z_i$,
and set $p_{c,k}\leftarrow \bar{z}_c$ for all $k=1,\dots,K$,followed by per-slot L2 normalization.
This yields a stable multi-prototype anchor at the beginning of fine-tuning.

During fine-tuning, given a minibatch of labeled foreground ROI features
$\mathcal{B}_c=\{z_i\}_{i=1}^{N_c}$ for class $c$ (each $z_i$ is L2-normalized),
we compute cosine similarities between samples and normalized prototypes:
\begin{equation}
s_{i,k} \;=\; z_i^{\top}\,\frac{p_{c,k}}{\lVert p_{c,k}\rVert_2}
\quad\text{for } k=1,\dots,K.
\end{equation}
Each sample is assigned to its slot $k_i^\ast=\arg\max_k s_{i,k}$,
yielding a submode-aware hard assignment to prototypes, without introducing
additional trainable weights beyond the $K$ prototypes.

Let $\mathcal{I}_{c,k} \!=\! \{\,i \mid k_i^\ast\!=\!k\,\}$ be the indices assigned to slot $k$.
If $\mathcal{I}_{c,k}\!\neq\!\emptyset$, we form the local mean
$\mu_{c,k} \!=\! \frac{1}{|\mathcal{I}_{c,k}|}\sum_{i\in\mathcal{I}_{c,k}} z_i$
and update the prototype with momentum $m\!\in\![0,1)$~\cite{tarvainen2017mean}:
\begin{equation}
p_{c,k} \;\leftarrow\; m\,p_{c,k} \;+\; (1-m)\,\mu_{c,k}.
\end{equation}
If no samples are assigned ($\mathcal{I}_{c,k}\!=\!\emptyset$), $p_{c,k}$ is kept unchanged.
After the update, prototypes are L2-normalized.
Compared to updating a fixed slot (e.g., slot~1) only, EMA allows
different slots to adapt to distinct intra-class submodes while keeping the update smooth and stable.

We apply EMA only on labeled foreground ROIs from the training data
within the fine-tuning stage. No statistics from the test set are used.
At inference time, the memory is frozen. The assignment costs $\mathcal{O}(N_c K D)$ for dot products and the update is $\mathcal{O}(N_c D)$.
With small $K$ (e.g., $K\!\le\!3$), the memory overhead is negligible.

\begin{algorithm}
\caption{PDA Prototype Memory Update (Fine-tuning)}
\label{alg:ema}
\begin{algorithmic}[1]
\Require Labeled batch $\{(\vect{f}_i, y_i)\}$; momentum $m$; projection $\vect{W}$; memory $\{\vect{p}_{c,k}\}_{c,k}$.
\For{each foreground class $c$ present in the batch}
  \State $\mathcal{B}_c \gets \{ i \mid y_i = c \}$
  \State \textbf{// normalize \& project}
  \For{$i \in \mathcal{B}_c$}
    \State $\tilde{\vect{f}}_i \gets \vect{f}_i / \|\vect{f}_i\|_2$;\quad $\vect{z}_i \gets \vect{W}\,\tilde{\vect{f}}_i$;\quad $\vect{z}_i \gets \vect{z}_i / \|\vect{z}_i\|_2$
  \EndFor
  \State \textbf{// hard routing to $K$ slots}
  \For{$i \in \mathcal{B}_c$}
    \State $k_i^\ast \gets \arg\max_{k=1..K}\ \vect{z}_i^\top \big(\vect{p}_{c,k}/\|\vect{p}_{c,k}\|_2\big)$
  \EndFor
  \State \textbf{// per-slot EMA}
  \For{$k=1$ \textbf{to} $K$}
    \State $\mathcal{I}_{c,k} \gets \{\, i\in\mathcal{B}_c \mid k_i^\ast = k \,\}$
    \If{$\mathcal{I}_{c,k} \neq \emptyset$ \textbf{ and } \texttt{FREEZE\_MEM} is False}
      \State $\boldsymbol{\mu}_{c,k} \gets \frac{1}{|\mathcal{I}_{c,k}|}\sum_{i\in\mathcal{I}_{c,k}} \vect{z}_i$
      \State $\vect{p}_{c,k} \leftarrow m\,\vect{p}_{c,k} + (1-m)\,\boldsymbol{\mu}_{c,k}$
      \State $\vect{p}_{c,k} \leftarrow \vect{p}_{c,k} / \|\vect{p}_{c,k}\|_2$
    \EndIf
  \EndFor
\EndFor
\State \textbf{return} updated memory $\{\vect{p}_{c,k}\}$
\end{algorithmic}
\end{algorithm}


\subsection{Best-of-$K$ Metric Scoring}
\label{subsec:bestofk}

Let $\mathbf{z}\in\mathbb{R}^{D}$ be the RoI feature after projection and L2 normalization,
and let $\{\mathbf{p}_{c,k}\}_{k=1}^{K}$ be the prototypes for class $c\in\{1,\dots,C\}$,
also L2-normalized as $\hat{\mathbf{p}}_{c,k}=\mathbf{p}_{c,k}/\lVert\mathbf{p}_{c,k}\rVert_2$.
We compute cosine similarities and take the best-of-$K$ score per class:
\begin{equation}
\label{eq:bestofk}
s_{c,k} \;=\; \mathbf{z}^{\top}\hat{\mathbf{p}}_{c,k},\qquad
s_c \;=\; \max_{k\in\{1,\ldots,K\}}\, s_{c,k},\qquad
\boldsymbol{s} \;=\; [s_1,\ldots,s_C] \in \mathbb{R}^{C}.
\end{equation}

We refine the similarity by first selecting,for each class $c$, its best prototype index
\begin{equation}
k_c^\star \;=\; \arg\max_{k\in\{1,\ldots,K\}} \; s_{c,k},
\qquad
\hat{\vect{p}}_{c}^\star \;=\; \hat{\vect{p}}_{c,k_c^\star}.
\end{equation}
Let $\mathcal{F}\!\in\!\mathbb{R}^{C_{\text{feat}}\times H\times W}$ be the RoI feature map (before pooling) that
produced $\vect{z}$. A lightweight aligner $A(\cdot\,;\,\hat{\vect{p}}_{c}^\star)$ predicts a 2D
offset field conditioned on the selected prototype, and warps $\mathcal{F}$ via grid-sampling to
obtain an aligned feature map $\tilde{\mathcal{F}}_c$.
We then average-pool and (optionally) project to get an aligned vector
$\tilde{\vect{z}}_c\in\mathbb{R}^{D}$ and recompute the cosine:
\begin{equation}
s_{c,k}^{\text{align}} \;=\; 
\frac{\tilde{\vect{z}}_c}{\|\tilde{\vect{z}}_c\|_2}^{\!\top}\,
\hat{\vect{p}}_{c,k},
\qquad
s_c \;\leftarrow\; \max_{k} \, s_{c,k}^{\text{align}}.
\end{equation}
If \texttt{USE\_ALIGN} is off, we use Eq.~\eqref{eq:bestofk} directly.
The rest of PDA logits and logit-level fusion remain unchanged.

With temperature $\tau>0$ and a learnable global scale $\sigma=\exp(\lambda)$,
PDA logits are
\begin{equation}
\label{eq:pda-logits}
\boldsymbol{z}^{\text{PDA}}
\;=\;
\sigma\;\operatorname{pad}_{\text{bg}}\!\big(\boldsymbol{s}/\tau\big) \in \mathbb{R}^{C+1},
\quad
\operatorname{pad}_{\text{bg}}(\boldsymbol{u}) \;=\; \big[u_1,\ldots,u_C,\; b_{\text{bg}}\big],
\end{equation}
where $b_{\text{bg}}\!\in\!\mathbb{R}$ is a learnable background bias to align PDA with the $(C{+}1)$-way classifier.

Following Fig.~\ref{fig:arch-fusion}, we fuse PDA, baseline, and PCB in the logit space:
\begin{equation}
\label{eq:fusion-vector}
\hat{\boldsymbol{z}}
\;=\;
\alpha\,\boldsymbol{z}^{\text{PDA}}
\;+\;
\beta \,\boldsymbol{z}^{\text{cls}}
\;+\;
\gamma\,\boldsymbol{z}^{\text{pcb}},
\end{equation}
where $\alpha,\beta,\gamma$ are fusion weights (set $\gamma{=}0$ if PCB is absent).
Final class scores are obtained via $\mathrm{softmax}(\hat{\boldsymbol{z}})$, and the box regressor is unchanged.

\begin{algorithm}
\caption{PDA Inference}
\label{alg:pda-min-align}
\begin{algorithmic}[1]
\State $\hat{\vect{f}} \gets \vect{f}/\|\vect{f}\|_2$;\quad $\vect{z}_0 \gets \vect{W}\,\hat{\vect{f}}$;\quad $\hat{\vect{z}}_0 \gets \vect{z}_0/\|\vect{z}_0\|_2$
\State $(c^\star,k^\star) \gets \arg\max_{c,k}\ \hat{\vect{z}}_0^{\top}\hat{\vect{p}}_{c,k}$
\State $\tilde{\mathcal{F}} \gets \mathrm{Align}\!\left(\mathcal{F},\,\hat{\vect{p}}_{c^\star,k^\star}\right)$
\State \textbf{// pooled \& projected feature for metric scoring}
\State $\vect{u} \gets \mathrm{GAP}(\tilde{\mathcal{F}})$;\quad
      $\vect{z} \gets \vect{W}\,\vect{u}$;\quad
      $\hat{\vect{z}} \gets \vect{z}/\|\vect{z}\|_2$
\For{$c = 1$ \textbf{to} $N$}
  \State $s_c \gets \max_{k=1..K}\ \hat{\vect{z}}^{\top}\hat{\vect{p}}_{c,k}$ \Comment{best-of-$K$ cosine}
\EndFor
\State $\boldsymbol{z}^{\mathrm{PDA}} \gets \sigma \cdot \big[\, s_1/\tau,\ldots,s_N/\tau,\ b_{\mathrm{bg}}\,\big]$ \Comment{pad background to $N{+}1$}
\State $\hat{\boldsymbol{z}} \gets \alpha\,\boldsymbol{z}^{\mathrm{PDA}} + \beta\,\boldsymbol{z}^{\mathrm{cls}} + \gamma\,\boldsymbol{z}^{\mathrm{pcb}}$ \Comment{set $\gamma{=}0$ if no PCB}
\State \textbf{return} $\mathrm{softmax}(\hat{\boldsymbol{z}})$
\end{algorithmic}
\end{algorithm}

\section{Experiments}
\label{sec:exp}

\subsection{Existing Benchmarks}
\label{subsec:bench}
We evaluate on the \textbf{PASCAL VOC} ~\cite{everingham2010pascal} benchmark using the \emph{standard three splits} for few-shot object detection. 
We report results under both \textbf{FSOD} (novel-only evaluation) and \textbf{GFSOD} (base+novel evaluation) protocols.
Unless otherwise specified, metrics include \texttt{AP}, \texttt{AP50}, and \texttt{AP75} over all evaluated classes; for GFSOD we additionally report \texttt{bAP} (base) and \texttt{nAP} (novel).
Each result is averaged over \textbf{10 pre-defined support draws} (fixed random seeds) for reproducibility.

\subsection{Evaluation Setting}
\label{subsec:eval}
We follow the common $K$-shot setup with $K\!\in\!\{1,2,3,5,10\}$ for each novel category, where the support set strictly contains \emph{exactly} $K$ labeled instances per novel class.
Unless stated otherwise:
(i) prototypes are initialized \emph{only} from the support set;
(ii) the prototype memory is \emph{frozen at inference};
(iii) we do \emph{not} use query/test statistics.
Settings that enable online EMA updates during fine-tuning are clearly marked as ablations (Sec.~\ref{subsec:ablation}).
For GFSOD, the detector is evaluated jointly on base and novel classes; the base head and regression branches remain unchanged.

\subsection{Implementation Details}
\label{subsec:impl}
\paragraph{Backbone and detector.}
We adopt \textbf{DeFRCN} with a \textbf{ResNet-101} backbone, following the official configuration. 
Gradient-Decoupled Layers (GDL) are placed as in the released code. 
RPN and the box regression head are kept unchanged.

\paragraph{PDA head.}
Given RoI features, the PDA head includes:
(i) a class-agnostic linear projection $\vect{W}$ (identity-initialized, trained end-to-end);
(ii) a best-of-$K$ prototype metric with temperature $\tau$ and a learnable global scale $\sigma{=}\exp(\lambda)$, plus a learnable background bias $b_{\mathrm{bg}}$;
(iii) an \emph{optional} prototype-conditioned RoI alignment (ProtoAlign).
EMA momentum $m\!\in\![0,1)$ controls online updates during fine-tuning; by default, \texttt{FREEZE\_MEM}$=$\texttt{True} (support-only prototypes at inference), and \texttt{FREEZE\_MEM}$=$\texttt{False} is reported as an ablation.

\paragraph{Training and fusion.}
We follow DeFRCN’s base-training and novel fine-tuning schedules.
For FSOD, the classifier is initialized from the standard surgery checkpoint; for GFSOD, the classification layer is expanded to include both base and novel classes.
Unless PCB is explicitly used, we set fusion weights $(\alpha,\beta,\gamma){=}(0.1,0.9,0.0)$; when PCB is present we tune $\gamma{>}0$ on the validation split while keeping $(\alpha,\beta)$ fixed.
Optimization uses SGD with base learning rate $1\!\times\!10^{-2}$ and the same batch size and training epochs as DeFRCN.
All other hyperparameters follow the DeFRCN defaults unless specified.

\subsection{Main Results}
\label{subsec:main}
\paragraph{FSOD on VOC.}
Table~\ref{tab:voc-fsod-novelsets} summarizes FSOD performance on VOC splits under 1/2/3/5/10-shot. 
PDA improves novel AP consistently over DeFRCN, with larger gains in extremely low-shot regimes (1–3 shots). 
Ablations suggest that the identity-initialized projection and best-of-K matching also contribute to the improvements.

\begin{table}[h]
  \centering
  \caption{VOC FSOD (novel-only) AP50 per novel set. }
  \label{tab:voc-fsod-novelsets}
  \begin{tabular}{lcccc}
    \toprule
    \textbf{Method} & \textbf{Shot} & \textbf{Novel Set 1} & \textbf{Novel Set 2} & \textbf{Novel Set 3} \\
    \midrule
    DeFRCN                  & 1  & 53.6 & 30.1 & 48.4 \\
    \;\; + PDA (ours)     & 1  & \textbf{56.9} & \textbf{32.8} & \textbf{51.5} \\
    \midrule
    DeFRCN                  & 2  & 57.5 & 38.1 & 50.9 \\
    \;\; + PDA (ours)     & 2  & \textbf{59.6} & \textbf{41.2} & \textbf{53.2} \\
    \midrule
    DeFRCN              & 3  & 61.5 & 47.0 & 52.3 \\
    \;\; + PDA (ours)     & 3  & \textbf{63.8} & \textbf{47.8} & \textbf{54.9} \\
    \midrule
    DeFRCN              & 5  & 64.1 & \textbf{53.3} & 54.9 \\
    \;\; + PDA (ours)     & 5  & \textbf{65.3} & 53.1 & \textbf{56.2} \\
    \midrule
    DeFRCN              & 10 & 60.8 & 47.9 & 57.4 \\
    \;\; + PDA (ours)     & 10 & \textbf{62.7} & \textbf{48.7} & \textbf{57.7} \\
    \bottomrule
  \end{tabular}
\end{table}

\subsection{Ablation Studies}
\label{subsec:ablation}

\paragraph{Prototype memory: freeze vs. EMA.}
Table~\ref{tab:ablation-mem} compares \texttt{FREEZE\_MEM}$=$\{True, False\}.
Enabling EMA during fine-tuning (\texttt{False}) offers slightly higher \texttt{nAP} in low-shot while keeping variance small; at inference, memory is always frozen.

\begin{table}[h]
  \centering
  \caption{Ablation on prototype memory in VOC FSOD (novel-only) — AP50 per novel set. }
  \label{tab:ablation-mem}
  \begin{tabular}{lcccc}
    \toprule
    \textbf{Method} & \textbf{Shot} & \textbf{Novel Set 1} & \textbf{Novel Set 2} & \textbf{Novel Set 3} \\
    \midrule
    DeFRCN                              & 1  & 53.6 & 30.1 & 48.4 \\
    \;\; + PDA (FREEZE\_MEM=True)       & 1  & 54.2 & 31.5 & 49.9 \\
    \;\; + PDA (EMA during FT)          & 1  & \textbf{56.9} & \textbf{32.8} & \textbf{51.5} \\
    \midrule
    DeFRCN                              & 2  & 57.5 & 38.1 & 50.9 \\
    \;\; + PDA (FREEZE\_MEM=True)       & 2  & 58.1 & 40.3 & 52.6 \\
    \;\; + PDA (EMA during FT)          & 2  & \textbf{59.6} & \textbf{41.2} & \textbf{53.2} \\
    \midrule
    DeFRCN                              & 3  & 61.5 & 47.0 & 52.3 \\
    \;\; + PDA (FREEZE\_MEM=True)       & 3  & 62.5 & 47.6 & 54.2 \\
    \;\; + PDA (EMA during FT)          & 3  & \textbf{63.8} & \textbf{47.8} & \textbf{54.9} \\
    \midrule
    DeFRCN                              & 5  & 64.1 & 53.3 & 54.9 \\
    \;\; + PDA (FREEZE\_MEM=True)       & 5  & 65.1 & \textbf{53.5} & 55.6 \\
    \;\; + PDA (EMA during FT)          & 5  & \textbf{65.3} & 53.1 & \textbf{56.2} \\
    \midrule
    DeFRCN                              & 10 & 60.8 & 47.9 & 57.4 \\
    \;\; + PDA (FREEZE\_MEM=True)       & 10 & 62.0 & \textbf{49.2} & \textbf{58.0} \\
    \;\; + PDA (EMA during FT)          & 10 & \textbf{62.7} & 48.7 & 57.7 \\
    \bottomrule
  \end{tabular}
\end{table}

Across shots, freezing the memory at inference is essential for protocol alignment and yields stable gains.
Allowing EMA updates during fine-tuning provides additional plasticity in extremely low-shot settings (1–3 shots) and slightly improves nAP without increasing variance; beyond 5 shots, the gap closes and the frozen variant may be marginally stronger on some splits.
This suggests that PDA’s benefit comes from a better metric geometry rather than memorization.

\section{Conclusion}
In this paper, we presented PDA, a prototype-driven adaptation head for few-shot object detection that complements the detector’s classifier with a class-agnostic metric branch. PDA builds support-only multi-prototype memories, uses an identity-initialized projection (with an optional prototype-conditioned RoI alignment), and fuses metric and detector logits with temperature and learnable scaling. On VOC FSOD, PDA consistently improves novel-class AP, especially at 1–3 shots.These results suggest the gains stem from a better metric geometry rather than memorization, while keeping overhead minimal.

\end{document}